\pdfoutput=1

\documentclass[11pt]{article}

\usepackage{acl}

\usepackage{times}
\usepackage{latexsym}
\usepackage{bm}
\usepackage{CJKutf8}

\usepackage[T1]{fontenc}

\usepackage[utf8]{inputenc}

\usepackage{microtype}

\usepackage{graphicx}
\usepackage{multirow}
\usepackage{amsmath}

\title{Exploring Automatic Evaluation Methods\\ based on a Decoder-based LLM for Text Generation}

\author{Tomohito Kasahara, Daisuke Kawahara \\
     Waseda University\\
     \texttt{\{tomo\_k@ruri.,dkw@\}waseda.jp} \\}

\begin{document}
\maketitle

\begin{abstract}
Automatic evaluation of text generation is essential for improving the accuracy of generation tasks.
In light of the current trend towards increasingly larger decoder-based language models, we investigate automatic evaluation methods based on such models for text generation.
This paper compares various methods, including tuning with encoder-based models and large language models under equal conditions, on two different tasks, machine translation evaluation and semantic textual similarity, in two languages, Japanese and English.
Experimental results show that compared to the tuned encoder-based models, the tuned decoder-based models perform poorly.
The analysis of the causes for this suggests that the decoder-based models focus on surface word sequences and do not capture meaning.
It is also revealed that in-context learning of very large decoder-based models such as ChatGPT makes it difficult to identify fine-grained semantic differences.
\end{abstract}


\section{Introduction}
\label{sec:intro}
Neural network-based text generation models are used in various natural language processing tasks, including machine translation, dialogue systems, and text summarization.
However, the outputs from these models are open-ended, and there is no single correct answer, making the evaluation of generations difficult.
Manual evaluation is often used due to its high accuracy but incurs significant temporal and financial costs.
Therefore, automatic evaluation is essential for the rapid development of text generation models.


Automatic evaluation methods for text generation, such as BLEU~\citep{papineni-etal-2002-bleu} and ROUGE~\citep{lin-2004-rouge}, have been based mainly on surface word overlaps between the generated text and the reference text.
In recent years, with the development of self-supervised models such as BERT~\citep{devlin-etal-2019-bert} and BART~\citep{lewis-etal-2020-bart}, more accurate automatic evaluation methods have been proposed. For example, BERTScore~\citep{zhang2019bertscore} uses word embeddings obtained by these models.
Such methods can be classified along two axes: whether the model used is an encoder-based, decoder-based, or encoder-decoder-based architecture of Transformer~\citep{vaswani2017attention}, and whether tuning is performed.
While encoder-based methods with tuning are reported to be highly accurate \citep{rei-etal-2020-comet}, in-context learning without tuning is the mainstream in decoder-based methods.


In recent years, self-supervised decoder-based models have become larger and larger, as seen in GPT-4~\citep{openai2023gpt4}, Megatron-Turing~\citep{smith2022using}, and PaLM~\citep{ Chowdhery2022PaLMSL}.
These decoder-based self-supervised large language models are referred to as \textbf{LLMs} in this paper.
However, encoder-based models have remained relatively smaller than decoder-based ones.


Based on the above situation, this paper compares various methods, including tuning with encoder-based models and LLMs under equal conditions, on two different tasks, machine translation evaluation and semantic textual similarity (STS), in two languages, Japanese and English.
The results revealed the following three observations.

\begin{enumerate}
    \item When a decoder-based model is tuned, the accuracy is proportional to the model size up to a certain model size, but it reaches a ceiling.
    \item Compared to tuned encoder-based models, tuned decoder-based models perform poorly.
    \item In-context learning of very large decoder-based models such as ChatGPT\footnote{\url{https://openai.com/chatgpt}} makes it difficult to identify fine-grained semantic differences.
\end{enumerate}

The analysis of the causes for the poor performance of the tuned decoder-based models suggests that they focus on surface word sequences and do not capture meaning.
Note that our study focuses on evaluation methods under the assumption that reference text is available.


\section{Related Work}
\label{sec:relatedwork}

Automatic evaluation of text generation mainly requires the text generated by a model and the reference text.
The classic automatic evaluation metrics, such as BLEU, ROUGE, METEOR~\citep{banerjee-lavie-2005-meteor}, and CIDEr~\citep{vedantam2015cider}, are based on the n-gram overlap between these two texts. 
The biggest disadvantage of these metrics is that they do not score well even when synonyms are included, as the n-grams must match exactly for a higher score.
TER~\citep{snover-etal-2006-study} and others that base their evaluation on edit distance have similar drawbacks.
METEOR aims to overcome this drawback by using a synonym dictionary, but it is unable to perform context-sensitive synonym evaluation.


Using embeddings derived from self-supervised models, synonyms can be judged to be similar based on their context.
BERTScore~\citep{zhang2019bertscore} is a method that embeds the generated text and the reference text respectively by an encoder-based model and calculates a score based on their similarity.
BARTScore~\citep{yuan2021bartscore} and T5Score~\citep{qin2022t5score} input the source text to the encoder and the target text to the decoder, and calculate a score based on the generation probability of the target text.
GPTScore~\citep{fu2023gptscore} calculates a score based on the generation probability of the target text by applying in-context learning~\citep{brown2020language} to an LLM.
G-Eval~\citep{liu2023gpteval} proposes a method to have an LLM generate scores directly.
In addition, \citet{chen2023exploring} show that directly generated scores are more accurate than generation probability-based ones when using LLMs.


Other evaluation methods increase accuracy by fine-tuning a self-supervised model using datasets consisting of text pairs and their similarity labels. 
Models trained on translation evaluation datasets include BLEURT~\citep{sellam-etal-2020-bleurt} and COMET~\citep{rei-etal-2020-comet}, while models trained on STS datasets include Sentence-BERT~\citep{reimers-gurevych-2019-sentence}.
There are also methods such as SimCSE~\citep{gao-etal-2021-simcse} that learn sentence embeddings by contrastive learning on natural language inference datasets and use them to calculate text pair similarity.
Most of these self-supervised methods use encoder-based models.
InstructScore~\citep{xu2023instructscore} is a method of fine-tuning LLaMA~\citep{touvron2023llama}.
However, \citet{xu2023instructscore}'s experiments did not involve tuned LLMs on the target datasets and did not compare them to encoder-based models under equal conditions.
In this study, we compare LLMs, which do not have bidirectional attention but larger model size, with encoder-based models, which have bidirectional attention but smaller model size, by tuning them under equal conditions.

\begin{table*}[t]
\small
\centering
\begin{tabular}{l|l|l|r||c|c|c|c}
\hline
Method & Model & Architecture & \multicolumn{1}{c||}{Size} & WMT20 & WMT21 & STS-B & SICK\\\hline
\hline 
\multicolumn{7}{l}{\textbf{No-Tuning Methods}} \\\hline
BLEU & - & - & - & 0.109 & 0.120 & 0.244 & 0.354 \\ 
Edit Distance & - & - & - & 0.345 & 0.340 & 0.089 & 0.278 \\
BERTScore & RoBERTa-large & Encoder & 355M & 0.306 & 0.294 & 0.405 & 0.455 \\
BARTScore CNN+Para & BART-large & Enc-Dec & 406M & 0.225 & 0.219 & 0.475 & 0.505 \\
OpenAI Embeddings & text-embedding-ada-002 & Encoder & ? & 0.184 & 0.181 & 0.655 & 0.627 \\
ChatGPT Zero-Shot & gpt-3.5-turbo & Decoder & ? & 0.113 & 0.097 & 0.669 & 0.622 \\
ChatGPT Few-Shot & gpt-3.5-turbo & Decoder & ? & 0.175 & 0.136 & 0.618 & 0.656 \\\hline

\multicolumn{7}{l}{\textbf{Tuning Methods (Not Target Dataset)}} \\\hline
BLEURT-20 & RemBERT & Encoder & 576M & 0.345 & 0.323 & 0.620 & 0.574 \\
InstructScore & LLaMA & Decoder & 6.7B & 0.439 & 0.345 & 0.471 & 0.526 \\\hline

\multicolumn{7}{l}{\textbf{Tuning Methods (Target Dataset)}} \\\hline
COMET (WMT21 MQM) & XLM-RoBERTa-large & Encoder & 560M & 0.506 & 0.362 & -- & -- \\
RoBERTa Fine-Tuning & RoBERTa-large & Encoder & 355M & \textbf{0.699} & \textbf{0.391} & \textbf{0.737} & \textbf{0.658} \\
\multirow{6}{*}{LLM LoRA-Tuning} & \multirow{6}{*}{Cerebras-GPT} & \multirow{6}{*}{Decoder} & 111M & 0.589 & 0.362 & 0.540 & 0.425 \\
& & & 256M & 0.634 & 0.378 & 0.585 & 0.462 \\
& & & 590M & 0.654 & 0.371 & 0.616 & 0.486 \\
& & & 1.3B & 0.663 & 0.383 & 0.625 & 0.483 \\
& & & 2.7B & 0.671 & 0.377 & 0.661 & 0.512 \\
& & & 6.7B & 0.665 & 0.370 & 0.681 & 0.530 \\\hline
\end{tabular}
\caption{Kendall's correlation coefficients between the predictions by the automatic evaluation metrics and the labels in the experiments in English.}
\label{tbl:corr_result_en}
\end{table*}

\section{Experimental Setup}
\label{sec:experimental_setup}

We compare various methods for text generation evaluation, including tuned encoder-based models and LLMs on equal conditions, on two different tasks, machine translation evaluation and STS, in two languages, Japanese and English.


\subsection{Datasets}
\label{ssec:dataset}

\subsubsection{Datasets in English}
\label{sssec:dataset_en}
For the experiments in English, we use WMT20~\citep{mathur-etal-2020-results} and WMT21~\citep{freitag-etal-2021-results} as the translation evaluation datasets, and STS-B~\citep{cer-etal-2017-semeval} and SICK~\citep{marelli-etal-2014-sick} as the datasets for STS.
WMT20 and WMT21 include human-translated texts, machine-translated texts, and their evaluation labels of Direct Assessment (DA) and Multidimensional Quality Metrics (MQM).
In our experiments, we adopted the MQM labels that were evaluated by experts and native speakers. Since only the Chinese-to-English translation task is labeled with MQM, we use its datasets (WMT20 MQM and WMT21 MQM).
STS and SICK consist of sentence pairs and their similarity labels.
Note that for WMT20 and WMT21, the datasets were not pre-separated into train, valid, and test, and we randomly split these datasets with a ratio of 8:1:1.


\subsubsection{Datasets in Japanese}
\label{sssec:dataset_ja}
The datasets used in the experiments in Japanese are the WMT20 English to Japanese translation task (WMT20 en-ja) and JSTS included in the Japanese General Language Understanding Evaluation (JGLUE)~\citep{kurihara-etal-2022-jglue} benchmark.
The WMT20 dataset includes human-translated texts, machine-translated texts, and their evaluation labels (Direct Assessment).
JSTS is an STS dataset for Japanese, consisting of sentence pairs and their similarity labels.
Note that WMT20 en-ja was randomly split at a ratio of train:valid:test=8:1:1 as in the English datasets.

\subsection{Tuning of LLMs}
\label{ssec:llm_lora}
For the method by LLM tuning, we performed LoRA-tuning of LLMs using datasets of text pairs and their evaluation or similarity labels.
We chose LoRA-tuning because it can achieve competitive accuracy with fine-tuning at a lower cost~\citep{hu2021lora}.


\subsubsection{Architecture and Input-Output Relationships}
\label{sssec:architecture}
\begin{figure}[t]
\centering
\includegraphics[width=1\linewidth]{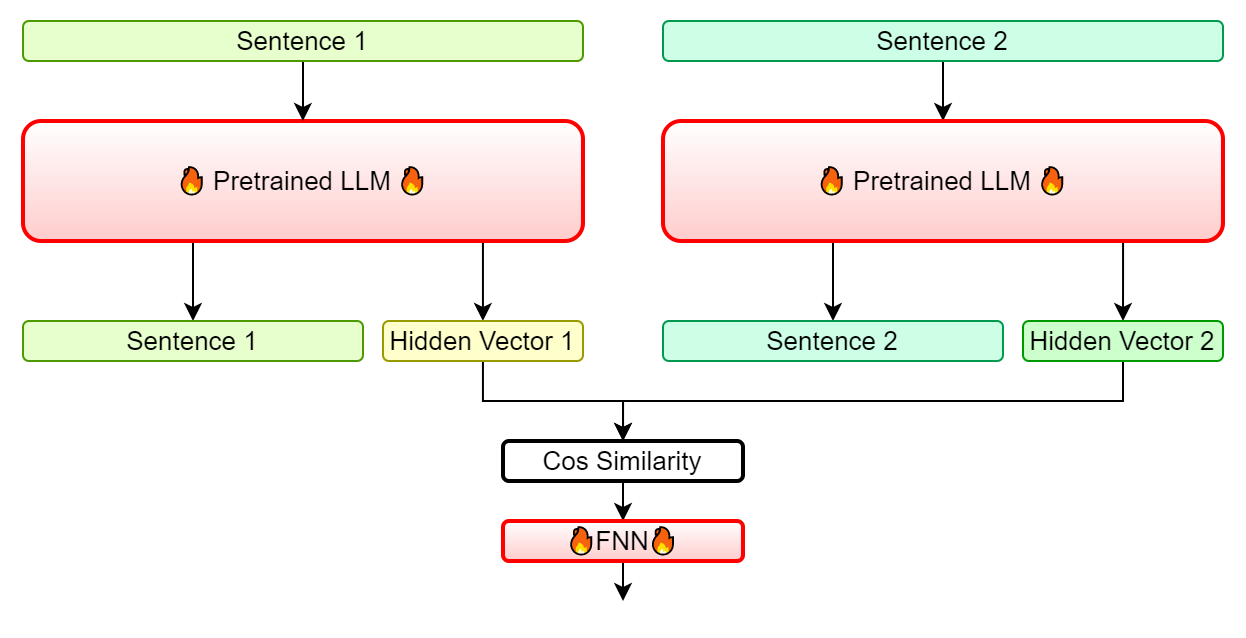}
\caption{The architecture and input-output overview of the LLM's tuning.}
\label{fig:model}
\end{figure}

The architecture and input-output relationship of the LLM's tuning are shown in Figure~\ref{fig:model}.
Given a text pair as an input to the model, their similarity value is returned as an output.
The following procedure is used to calculate the similarity.


\begin{enumerate}
    \item Feed each text of a text pair into an LLM.
    \item Obtain the embedding corresponding to the token at the end of each text (the preceding token of the EOS token).
    \item Calculate the cosine similarity between the two embeddings.
    \item Pass the cosine similarity to a 1-layer FNN and regard its output as the similarity of the text pair.
\end{enumerate}


The FNN layer is used to convert the cosine similarity values into a label distribution of the dataset.
Based on the results of our preliminary experiments, we decided to use the embedding of the token at the end of a text instead of the special EOS token.


\begin{table*}[h]
\small
\centering
\begin{tabular}{l|l|l|r||c|c}
\hline
Method & Model & Architecture & \multicolumn{1}{c||}{Size} & WMT20 & JSTS\\\hline
\hline 
\multicolumn{5}{l}{\textbf{No-Tuning Methods}} \\\hline
BLEU & - & - & - & 0.226 & 0.353 \\ 
Edit Distance & - & - & - & 0.242 & 0.321 \\
BERTScore & Waseda RoBERTa-large & Encoder & 337M & 0.319 & 0.558 \\
OpenAI Embeddings & text-embedding-ada-002 & Encoder & ? & 0.237 & 0.611 \\
ChatGPT Zero-Shot & gpt-3.5-turbo & Decoder & ? & 0.187 & 0.709 \\
ChatGPT Few-Shot & gpt-3.5-turbo & Decoder & ? & 0.205 & 0.690 \\\hline

\multicolumn{5}{l}{\textbf{Tuning Methods (Not Target Dataset)}} \\\hline
BLEURT-20 & RemBERT & Encoder & 576M & 0.315 & 0.569 \\\hline

\multicolumn{5}{l}{\textbf{Tuning Methods (Target Dataset)}} \\\hline
RoBERTa Fine-Tuning & Waseda RoBERTa-large& Encoder & 337M & \textbf{0.396} & \textbf{0.729} \\
\multirow{5}{*}{LLM LoRA-Tuning} & \multirow{5}{*}{Rinna-gpt} & \multirow{5}{*}{Decoder} & 37M & 0.342 & 0.600 \\
& & & 110M & 0.378 & 0.644 \\
& & & 336M & \textbf{0.396} & 0.677 \\
& & & 1.3B & 0.370 & 0.659 \\
& & & 3.6B & 0.380 & 0.687 \\\hline
\end{tabular}
\caption{Kendall's correlation coefficients between the predictions by the automatic evaluation metrics and the labels in the experiments in Japanese.}
\label{tbl:corr_result_ja}
\end{table*}

\subsubsection{Training Method}
\label{sssec:train_method}
The gold labels (similarity values) in the dataset are normalized between 0 and 1 in advance.
We calculate the similarity of a text pair using the procedure described in Section~\ref{sssec:architecture}.
Next, only the parameters newly added to the model (including the parameters of the FNN) are updated based on the mean squared error between the predictions and the gold labels.
Furthermore, the initial values of the FNN are set to 1 for weight and 0 for bias.
We employ LoRA-tuning as the tuning method of the LLM for its high performance.


For experiments in English, we use the Cerebras-GPT models\footnote{\url{https://huggingface.co/cerebras}} with parameter sizes ranging from 111M to 6.7B.
These models are tuned on WMT20 MQM for the translation evaluation task and on STS-B for the STS tasks, respectively.
In other words, the models trained with WMT20 MQM are evaluated on WMT20 MQM and WMT21 MQM, and the models trained with STS-B are evaluated on STS-B and SICK.


For experiments in Japanese, we use the GPT-2 and GPT-NeoX models developed by rinna\footnote{\url{https://huggingface.co/rinna}}, ranging from the 37M model to the 3.6B model.
We trained models on each of the two datasets in Section~\ref{sssec:dataset_ja}.


\subsection{Baselines}
\label{ssec:baseline}
For comparison, we adopt the following baselines: BLEU, character edit distance, fine-tuned RoBERTa-large~\citep{liu2019roberta}, BERTScore\footnote{\url{https://github.com/Tiiiger/bert_score }}, BARTScore\footnote{\url{https://github.com/neulab/BARTScore}}, OpenAI Embeddings~\cite{neelakantan2022text}, in-context learning of ChatGPT (gpt-3.5-turbo), BLEURT\footnote{\url{https://github.com/google-research/bleurt}}, COMET\footnote{\url{https://unbabel.github.io/COMET}} and InstructScore\footnote{\url{https://github.com/xu1998hz/SEScore3}}.
For fine-tuned RoBERTa, as described in Section~\ref{sssec:train_method}, we trained models on WMT20 MQM and STS-B for the English experiments and on the two datasets shown in Section~\ref{sssec:dataset_ja} for the Japanese experiments, respectively. 
For BERTScore, the training data is used to select the best output layer to obtain the embeddings.
For OpenAI Embeddings, the scores are the cosine similarity of the obtained embeddings.
The prompt used in ChatGPT's in-context learning is shown in Appendix~\ref{sec:prompt}.
We also had a preliminary experiment with in-context learning of Cerebras-GPT as well as ChatGPT, but were unable to generate scores successfully.
It is assumed that the model size of few billion is too small for in-context learning.
We do not tune BLEURT, but instead use BLEURT-20~\citep{pu-etal-2021-learning}, which is trained in multiple languages.
For COMET, we use the model trained on WMT21 MQM. 
We do not apply COMET to the STS datasets because COMET is a metric for automatic translation evaluation and requires three inputs: pre-translated text, human-translated text, and machine-translated text.
Our hyperparameters for training are shown in Appendix~\ref{sec:hyperparameter}.

Note that BARTScore, COMET, and InstructScore, only support English and hence are not used for experiments in Japanese.



\begin{table*}[t]
\small
\centering
\begin{tabular}{l|r||c|c|c|c|c|c|c|c}
\hline
Model & \multicolumn{1}{c||}{Size} & \multicolumn{4}{c|}{BLEU} & \multicolumn{4}{c}{Edit Distance}\\
&& WMT20 & WMT21 & STS-B & SICK & WMT20 & WMT21 & STS-B & SICK \\\hline
\hline 

RoBERTa-large & 355M & 0.126 & 0.127 & 0.237 & 0.363 & 0.394 & 0.511 & 0.046 & 0.262 \\\hline
\multirow{6}{*}{Cerebras-GPT} & 111M & 0.213 & 0.212 & 0.281 & 0.553 & 0.491 & 0.612 & 0.077 & 0.491 \\
& 256M & 0.192 & 0.216 & 0.292 & 0.553 & 0.455 & 0.615 & 0.081 & 0.486 \\
& 590M & 0.187 & 0.211 & 0.268 & 0.559 & 0.432 & 0.583 & 0.087 & 0.487 \\
& 1.3B & 0.175 & 0.225 & 0.277 & 0.545 & 0.425 & 0.574 & 0.096 & 0.483 \\
& 2.7B & 0.178 & 0.231 & 0.263 & 0.549 & 0.428 & 0.567 & 0.058 & 0.478 \\
& 6.7B & 0.181 & 0.205 & 0.259 & 0.552 & 0.441 & 0.522 & 0.068 & 0.472 \\\hline
\end{tabular}
\caption{Kendall's correlations between the metrics based on superficial word sequences and the predictions by models with tuning in the experiments in English.}
\label{tbl:analyze_en}
\end{table*}

\section{Experimental Results and Analysis}
\label{sec:experimental_results}

\subsection{Main Results}
Kendall's correlation coefficients between the predictions by the automatic evaluation metrics and the gold labels in English and Japanese are shown in Tables~\ref{tbl:corr_result_en} and \ref{tbl:corr_result_ja}, respectively.
For all datasets in both languages, RoBERTa-large with fine-tuning achieved the highest accuracy.
For LoRA-tuned LLMs, there is a tendency for the accuracy to be proportional to the model size up to a certain model size, but it reaches a ceiling.
Also, even models with overwhelmingly larger parameter sizes than RoBERTa-large showed low accuracy.
For ChatGPT's in-context learning, the accuracy on the STS datasets was comparable to that of the tuning-based methods, but its accuracy on the translation evaluation datasets was low.
Note that most of the p-values were very close to 0.


\subsection{Analysis of Why Tuned LLMs are Inferior}
\label{ssec:llm_lora_analysis}
From Tables~\ref{tbl:corr_result_en} and \ref{tbl:corr_result_ja}, we observe that LoRA-tuned LLMs, which have by far a larger number of parameters than RoBERTa-large, are inferior in terms of performance.
We analyze the causes of this from the experimental results in English.

The most significant difference between the two models is that RoBERTa, an encoder-based model, has bidirectional attention, while an LLM has unidirectional attention.
Here, we hypothesized that unidirectional attention focuses more on surface word sequences as opposed to bidirectional attention.
To confirm this hypothesis, we calculated the correlations of the predictions of RoBERTa and LLMs to BLEU and character edit distance, which are the metrics based on superficial word sequences.
The results are shown in Table~\ref{tbl:analyze_en}.
As hypothesized, the results show that the correlations to both BLEU and edit distance are stronger for LLMs than the encoder-based model.
The fact that the correlation decreases as the model size increases in LLMs suggests that the larger the model size, the better the prediction is able to capture not only the surface word sequences but also the meaning of the text. 
However, even with a model size of 6.7B, the LLM is still not as accurate as RoBERTa.


\subsection{Analysis of the Inability of ChatGPT's In-context Learning}
\label{ssec:chatgpt_analysis}
While ChatGPT's in-context learning showed high accuracy on the STS datasets, it did not perform well on the translation evaluation datasets.
We analyze the causes of this from the experimental results in English.

\begin{figure}[t]
\centering
\small
\includegraphics[width=1\linewidth]{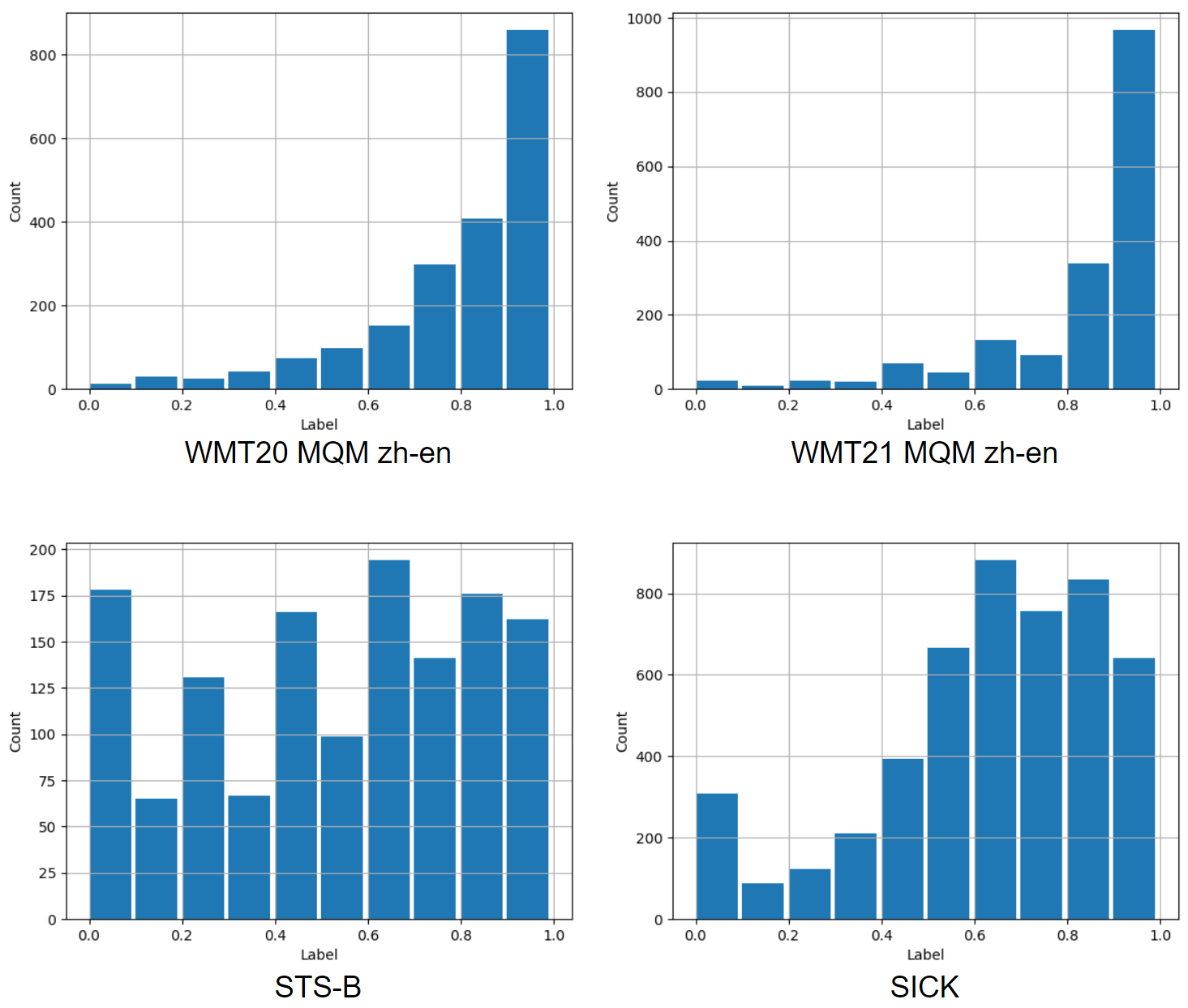}
\caption{Label distribution of the test datasets used in the English experiments.}
\label{fig:label_distribution}
\end{figure}

In our experiments, the prompts were created to score on a scale of 0 to 100. However, in the output scores, there were many cases where the last digit was 0 or 5 in both zero-shot and few-shot settings.
Also, as shown in Figure~\ref{fig:label_distribution}, the label distributions of the translation evaluation datasets are skewed between 0.9 and 1.0, compared to the STS datasets, which have gently sloping distributions.
Therefore, most of the predictions in the translation evaluation datasets are 95, etc., and this is thought to have caused the accuracy drop.
Thus, it is clear that ChatGPT's in-context learning has difficulty in identifying fine-grained semantic differences.


\section{Conclusion}
\label{sec:conclusion}
In this paper, we compared various automatic evaluation methods for text generation in two languages, Japanese and English.
We showed that fine-tuned encoder-based models are the strongest when training data is available, and in-context learning of ChatGPT is equally accurate when the variance of scores is large.
Our analysis also revealed that tuned LLMs are less accurate than tuned encoder-based models because of their focus on surface word sequences.


\section*{Limitations}
Our experiments assume the presence of a training dataset.
If no dataset for training exists, refer to the results without the \textbf{Tuning Method (Target Dataset)} to compare the metrics in Tables~\ref{tbl:corr_result_en} and \ref{tbl:corr_result_ja}.

\section*{Acknowledgements}
This work was supported by a joint research grant from LINE Corporation. 

\bibliography{anthology,custom}
\bibliographystyle{acl_natbib}

\clearpage
\appendix
\onecolumn

\section{Prompt Used in Experiments with ChatGPT}
\label{sec:prompt}
The following text is an example of the prompt used in our experiments with ChatGPT, which was created by referring to the prompt used in \citet{chen2023exploring}'s experiments.


\begin{figure}[h]
\centering
\includegraphics[width=0.8\linewidth]{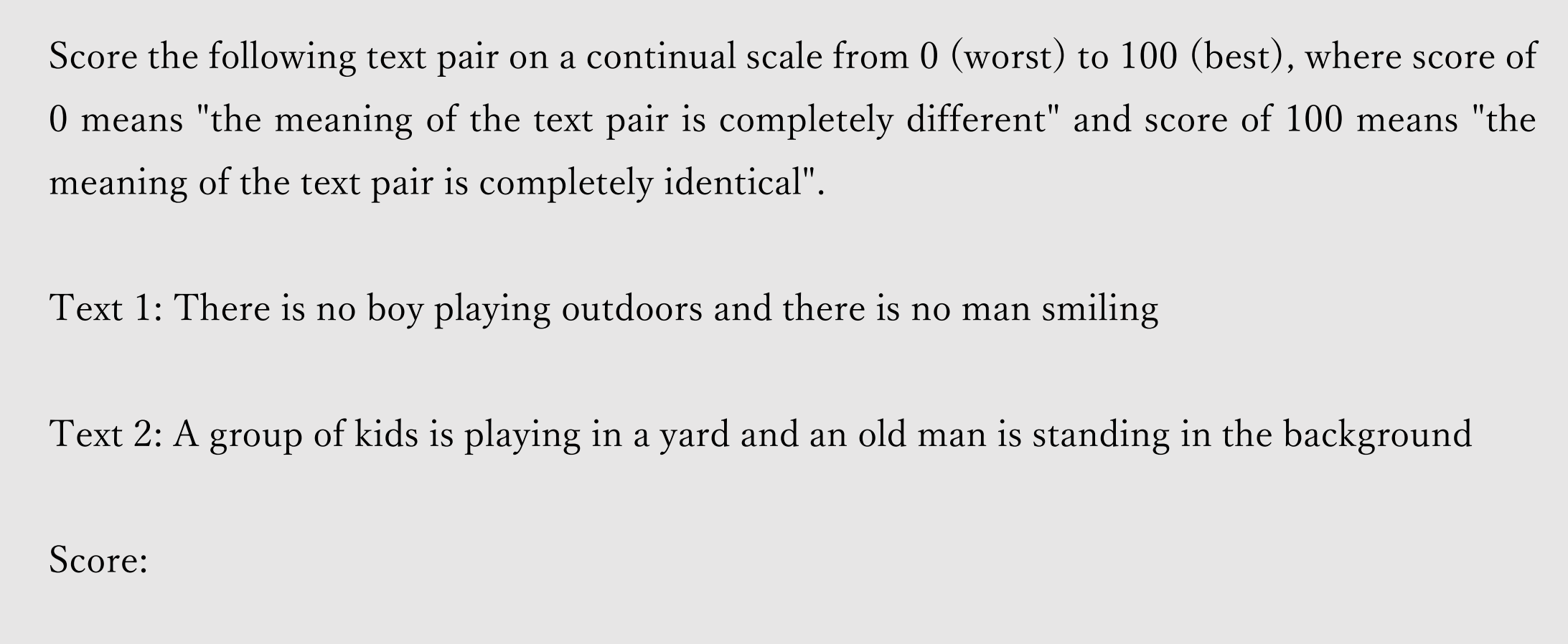}
\label{fig:prompt_example}
\end{figure}

\section{Hyperparameters}
\label{sec:hyperparameter}
The hyperparameters that we used for training models in our experiments are shown in Table~\ref{tbl:hyperparameter}.
Note that the GPU used in our experiments is the NVIDIA A100 SXM4 GPU with a GPU memory size of 40 GB.


\begin{table}[h]
\centering
\begin{tabular}{l||cc}
\hline
Hyperparameters & \begin{tabular}{c}RoBERTa\\Fine-Tuning\end{tabular} & \begin{tabular}{c}LLM\\LoRA-Tuning\end{tabular} \\\hline\hline
Learning Rate & 2e-5 & 2e-4, 1e-4, 5e-5, 1e-5\\
Epoch Num & 10 & 10\\
LoRA Dim & - & 4 \\
LoRA Alpha & - & 32 \\
LoRA Dropout & - & 0.1 \\
\hline
\end{tabular}
\caption{Hyperparameters for training in the experiments.}
\label{tbl:hyperparameter}
\end{table}

\end{document}